\documentclass[letterpaper,journal]{IEEEtran}
\usepackage[table,xcdraw]{xcolor} % 用于表格着色
\usepackage{amsmath,amsfonts}
\usepackage{algorithmic}
\usepackage{array}
\usepackage{cite}
\usepackage{pdfpages}
\usepackage{tabularx}
\usepackage{threeparttable}
\usepackage{multirow}
\usepackage[caption=false,font=normalsize,labelfont=sf,textfont=sf]{subfig}
\usepackage{textcomp}
\usepackage{stfloats}
\usepackage{url}
\usepackage{verbatim}
\usepackage{graphicx}
\usepackage{hyperref}
\usepackage{makecell}
\usepackage{balance}
\usepackage{soul}
\usepackage{times}
\usepackage{epsfig}
\usepackage{amssymb}
\usepackage{booktabs}
\usepackage{bbding}
\usepackage[normalem]{ulem}
\useunder{\uline}{\ul}{}
\usepackage{float}
\usepackage{bm}
\usepackage{pifont}
\hypersetup{
    hidelinks,
    colorlinks=true,
    allcolors=blue,
    pdfstartview=Fit,
    breaklinks=true
}

\newcommand{\Rmnum}[1]

\def\BibTeX{{\rm B\kern-.05em{\sc i\kern-.025em b}\kern-.08em
    T\kern-.1667em\lower.7ex\hbox{E}\kern-.125emX}}

\title{VM-BHINet:Vision Mamba Bimanual Hand Interaction Network for 3D Interacting Hand Mesh Recovery From a Single RGB Image}
\author{
  Han Bi, Ge Yu, Yu He, Wenzhuo Liu and Zijie Zheng
  \thanks{Corresponding author: Yu He. }
  \thanks{
    Han Bi and Zijie Zheng are with the college of Aeronautics and Astronautics, University of Chinese Academy of Sciences, Beijing, China. 

    Ge Yu, Yu He are with Key Laboratory of Space Utilization, Technology and Engineering Center for Space Utilization, Chinese Academy of Sciences, Beijing, China.

    Wenzhuo Liu is with the Faculty of Marine Science and Technology, Beijing Institute of Technology, Zhuhai, China.   
  }
}
\begin{document}
\maketitle

\begin{abstract}
Understanding bimanual hand interactions is essential for realistic 3D pose and shape reconstruction. However, existing methods struggle with occlusions, ambiguous appearances, and computational inefficiencies. To address these challenges, we propose Vision Mamba Bimanual Hand Interaction Network (VM-BHINet), introducing state space models (SSMs) into hand reconstruction to enhance interaction modeling while improving computational efficiency. The core component, Vision Mamba Interaction Feature Extraction Block (VM-IFEBlock), combines SSMs with local and global feature operations, enabling deep understanding of hand interactions. Experiments on the InterHand2.6M dataset show that VM-BHINet reduces Mean per-joint position error (MPJPE) and Mean per-vertex position error (MPVPE) by 2-3$\%$, significantly surpassing state-of-the-art methods.
\end{abstract}


\section{Introduction}
Bimanual interactions are fundamental in human activities, such as collaborative tasks, emotion expression, and intention communication. Understanding these interactions is crucial for applications in augmented reality (AR)/ virtual reality (VR)\cite{han2020megatrack, singh2024advancing}, human-computer interaction (HCI), and social signal understanding\cite{joo2019towards,gong2023sifdrivenet,liu2025mmtl, ng2021body2hands,liu2024glmdrivenet,liu2024fmdnet,gong2022multi,liu2025umd}. Accurate modeling and real-time reconstruction of bimanual interactions not only enhance system responsiveness but also improve the user experience by enabling natural and intuitive interactions.

In recent years, significant progress has been made in 3D hand pose estimation using monocular RGB images. Unlike earlier approaches that relied on multi-view setups\cite{ballan2012motion, han2020megatrack}, depth sensors\cite{mueller2019real, oikonomidis2012tracking, tzionas2016capturing, kyriazis2014scalable, taylor2016efficient}, or auxiliary hardware, monocular methods offer practical advantages, reducing deployment costs and hardware requirements. However, these methods face unique challenges, particularly in scenarios involving bimanual interactions. Severe occlusions, ambiguous hand appearances, and dynamic hand movements introduce significant complexity. Furthermore, existing methods often fail to model the intricate relationships between two interacting hands, leading to inaccurate and unnatural reconstructions. For example, during close interactions, occlusions can obscure large portions of the hands, and self-similar appearances can result in hand misassignments, significantly degrading the quality of the reconstruction.

\begin{figure}
\centering
\includegraphics[width=0.49\textwidth]{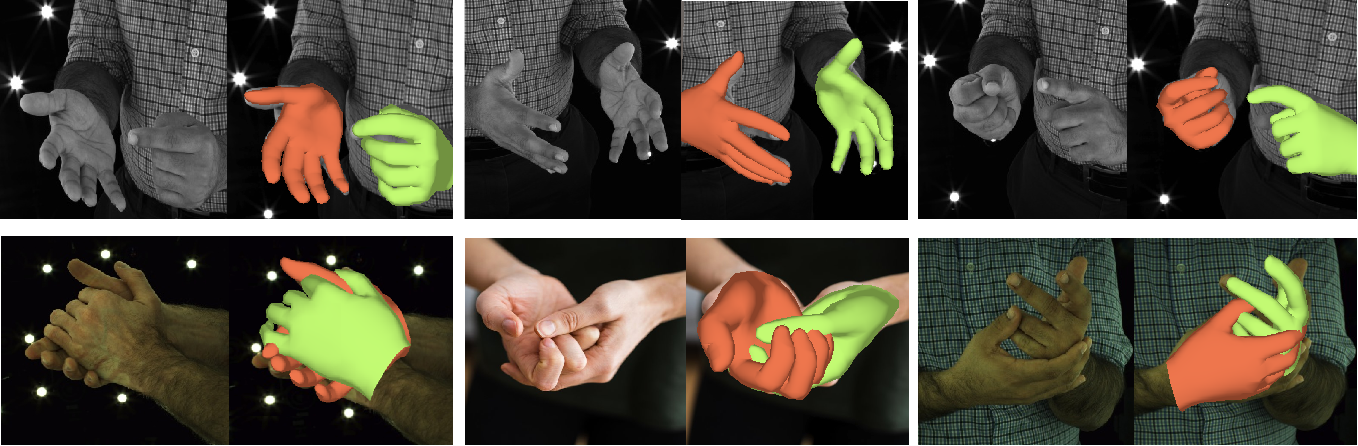}
\caption{Visual results of the VM-BHINet. VM-BHINet achieves remarkable visual performance in various hand poses.}
\vspace{-0.5cm}
\label{fig1}
\end{figure}

To address these limitations, we propose the Vision Mamba Bimanual Hand Interaction Network (VM-BHINet). This framework is the first to integrate state space models (SSMs) into 3D interacting hand reconstruction, enabling dynamic feature modeling and improving computational efficiency. The core component of VM-BHINet, the Vision Mamba Interaction Feature Extraction Block (VM-IFEBlock), combines local convolution operations with global feature representations to enhance the understanding of complex hand interactions. By leveraging state space models, the model captures long-range dependencies and dynamic interactions between hands, effectively addressing challenges like occlusions and appearance ambiguities. As shown in Figure \ref{fig1}, VM-BHINet achieves remarkable visual performance in various hand poses.

The primary contributions of this work are as follows:
\begin{itemize}
    \item[$\bullet$]We propose Vision Mamba Bimanual Hand Interaction Network (VM-BHINet), the first framework to incorporate state space models for 3D interacting hand reconstruction, significantly enhancing accuracy and computational efficiency.
    \item[$\bullet$]We introduce the Vision Mamba Interaction Feature Extraction Block (VM-IFEBlock), which integrates local and global modeling to effectively capture complex hand interactions.
    \item[$\bullet$]We perform comprehensive experiments on the InterHand2.6M dataset\cite{moon2020interhand2, moon2022neuralannot}, demonstrating that VM-BHINet reduces Mean per-joint position error (MPJPE) and Mean per-vertex position error (MPVPE) by 2-3$\%$, outperforming state-of-the-art methods.
\end{itemize}

\section{Related Work}
\label{sec:Related Work}

\subsection{3D interacting hand mesh recovery}
Research on 3D interacting hand mesh recovery has evolved through several paradigms, primarily categorized into Optimization-based Paradigm and Regression-based Paradigm, which further includes Model-based and Model-free approaches.

\subsubsection{Optimization-based Paradigm.}  

Early studies predominantly relied on optimization-based methods, where 3D hand models were fitted to geometric evidence derived from RGB-D sequences~\cite{oikonomidis2012tracking}, hand segmentation maps~\cite{mueller2019real}, and dense matching maps~\cite{wang2020rgb2hands}. These methods aimed to iteratively minimize the discrepancy between observed visual cues and reconstructed hand geometries. While effective under controlled scenarios, these methods often suffer from heavy computational overhead, initialization sensitivity, and vulnerability to local minima, limiting their applicability in real-time and dynamic interaction scenarios.

\subsubsection{Regression-based Paradigm.}  
Recently, regression-based methods have gained prominence due to their ability to directly predict hand pose and shape parameters from image inputs. These methods are broadly divided into Model-based and Model-free approaches.

\textbf{Model-based approaches.} 
Model-based methods primarily rely on parametric hand models, with MANO~\cite{romero2022embodied} being the most widely adopted. MANO represents a hand mesh using 778 vertices, 1538 faces, and 16 joints, enabling integration with deep learning architectures for 3D mesh reconstruction. Rong et al.~\cite{rong2021monocular} introduced a two-stage framework to minimize collisions in interacting hand meshes. Zhang et al.~\cite{zhang2021interacting} improved reconstruction accuracy using pose-aware and context-aware attention mechanisms. Kwon et al.~\cite{kwon2021h2o} proposed the H2O dataset and a graph-based network tailored for hand-object interactions. Li et al.~\cite{li2022interacting} leveraged attention-based modules for enhanced dual-hand reconstruction, while Hampali et al.~\cite{hampali2022keypoint} and Di et al.~\cite{di2022lwa} focused on lightweight architectures optimized for occlusion handling and computational efficiency. However, despite their structured representation and precision, model-based methods often struggle with initialization sensitivity, occlusion handling, and the computational complexity associated with mapping RGB images to hand model parameters.

\textbf{Model-free approaches.} 
To overcome the inherent limitations of model-based methods, model-free approaches have emerged as an alternative, directly predicting 3D hand structures without relying on parametric models~\cite{kim2021end, lin2021two, fan2021learning, meng20223d, cui2024textnerf, cheng2021handfoldingnet,tan2025graph, moon2022neuralannot}. Moon et al.~\cite{moon2020interhand2, moon2022neuralannot} introduced the InterHand2.6M dataset and the baseline network InterNet, which estimates 3D heatmaps for joint locations, improving robustness and interpretability. InterNet excels at handling single-hand and interacting-hand poses while predicting handedness and relative depth. Inspired by this, Kim et al.~\cite{kim2021end} used a GAN-based discriminator to improve pose realism, while Lin et al.~\cite{lin2021two} proposed a method for global hand pose estimation from monocular images. Fan et al.~\cite{fan2021learning} and Meng et al.~\cite{meng20223d} focused on resolving ambiguities arising from occlusions and interaction complexities. Cheng et al.~\cite{cheng2021handfoldingnet} introduced HandFoldingNet, a lightweight decoder-based network capable of achieving high accuracy with reduced computational costs. Overall, model-free methods offer enhanced flexibility, lightweight architectures, and improved real-time performance. However, challenges remain in capturing fine-grained interaction dependencies and maintaining robustness under severe occlusion scenarios.

In summary, optimization-based methods established the foundation for 3D hand reconstruction. Regression-based approaches, covering both model-based and model-free methods, have since demonstrated significant improvements. Nevertheless, current 3D interacting hand reconstruction methods still have limitations when handling scenes with close hand interactions. These limitations are mainly reflected in the following aspects: first, the insufficient capture of interaction features between the left and right hands, which leads to lower reconstruction accuracy in complex dynamic hand gesture interaction scenarios; second, the complex parameterization design of the model, which increases the difficulty of training and inference; and third, the high computational resource requirements, which limit the practical application potential of these methods. To address these issues, we propose a new method that improves reconstruction accuracy in complex interaction scenarios and enhances the model's practical application potential by optimizing feature extraction and model structure.

\begin{figure*}[h]
    \centering
    \includegraphics[width=0.98\textwidth]{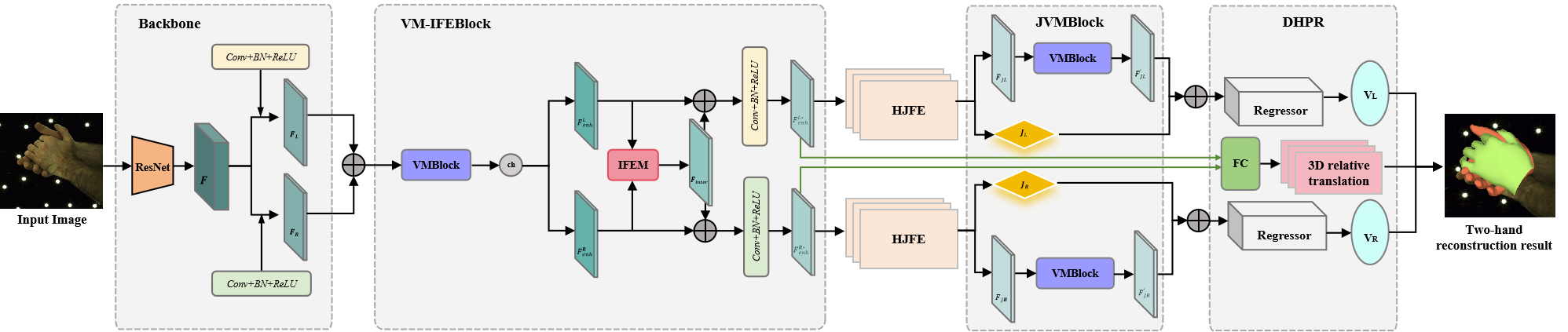}
    \caption{Our proposed Vision Mamba Bimanual Hand Interaction Network (VM-BHINet). It consists of five main components: the Backbone, the Vision Mamba Interaction Feature Extraction Block (VM-IFEBlock), Hand Joint Feature Extractor (HJFE), Joint Vision Mamba Block (JVMBlock), and the Dual Hand Parameter Regressor (DHPR).}
    \label{fig2}
\end{figure*}

\subsection{State Space Models (SSMs)}
\textbf{State Space Models (SSMs)} provide a mathematical framework for modeling dynamic systems using state variables and state transition equations. Initially applied in \textbf{Kalman Filtering}~\cite{kalman1960new}, State Space Models (SSMs) effectively estimate system states in noisy environments and have been widely used in signal processing, control systems, and robot navigation. Their ability to capture temporal dependencies and model dynamic behaviors makes them particularly useful for tasks requiring sequential data interpretation.

With recent advancements in deep learning and computer vision, State Space Models (SSMs) have been adapted to address more complex tasks~\cite{li2024mipd, huang2024mfe, zhang2023oblique, wang2023path, gan2024segmentation, shi2023bssnet}. Notable examples include the development of the \textbf{Structured State Space Sequence (S4) model}~\cite{gu2021efficiently, gu2021combining}, which enhances long-range dependency modeling. More recently, \textbf{Mamba}~\cite{gu2023mamba} has emerged as an optimized variant of State Space Models (SSMs), introducing dynamic weights and global receptive fields to handle long-sequence dependencies efficiently. Compared to traditional transformer-based architectures, which suffer from quadratic complexity in self-attention, Mamba achieves higher efficiency and scalability while maintaining competitive accuracy.

Mamba has demonstrated significant success in tasks such as image segmentation, object detection, and 3D reconstruction. However, its application in modeling hand interactions, particularly in 3D interacting hand reconstruction, remains largely unexplored. In this work, we introduce Vision Mamba Bimanual Hand Interaction Network (VM-BHINet), the first framework to integrate Mamba into the domain of 3D interacting hand reconstruction. By leveraging the dynamic modeling capabilities of State Space Models and the efficiency of Mamba, our approach addresses the challenges posed by occlusions, ambiguous hand poses, and computational overhead.

In conclusion, the integration of State Space Models, particularly the Mamba variant, presents a promising avenue for addressing the limitations of existing methods in 3D hand reconstruction. This work represents a significant step toward bridging the gap between dynamic sequence modeling and real-time hand interaction reconstruction.

\section{Methods}

In this section, we present the overall architecture of our proposed Vision Mamba Bimanual Hand Interaction Network (VM-BHINet), along with a detailed description of the structure and functionality of each individual module. As illustrated in Figure \ref{fig2}, VM-BHINet consists of five main components: the Backbone, the Vision Mamba Interaction Feature Extraction Block (VM-IFEBlock), Hand Joint Feature Extractor (HJFE), Joint Vision Mamba Block (JVMBlock), and Dual Hand Parameter Regressor (DHPR). Each component plays a vital role in processing and fine-tuning the input data, progressively enhancing the accuracy and robustness of the model's output, which includes intricate 3D hand mesh reconstructions and interaction details.

Our structure first obtains left and right hand features \( F_L \) and \( F_R \) through a backbone. These features are then concatenated and fed into the VM-IFEBlock. In this block, the concatenated features \( F_{concat} \) are initially processed for enhancement through the VMBlock. The enhanced features \( F_{enh} \) are then chunked again into left and right hand features \( F_{enh}^{L} \) and \( F_{enh}^{R} \) along the channel dimension. Following this, the Interaction Feature Extraction Module (IFEM) utilizes non-local attention mechanisms to extract deep-level interactive features \( F_{inter} \) between the hands. These interactive features \( F_{inter} \) are fused with the original features through subsequent convolutional layers to obtain \( F_{enh}^{L*} \) and \( F_{enh}^{R*} \), further enhancing the richness and discriminative power of the feature representation. This integration allows the fused features to provide more accurate and comprehensive data representation, thereby improving the model's performance, generalization ability, and robustness. Following this module, the Hand Joint Feature Extractor (HJFE) first estimates the 2.5D joint coordinates for each hand from the enhanced hand features using a convolutional layer. Then, by applying the soft-argmax operation to the 2.5D heatmap, it obtains the precise joint coordinates \( J_L \) and \( J_R \) for each hand and extracts the corresponding joint features (\( F_{JL} \) and \( F_{JR} \)). These provide the essential positional information and contextual features required for accurate hand pose estimation and 3D mesh reconstruction, forming the foundation for precise and natural hand reconstruction and interaction. The following JVMBlock processes the joint features of each hand through the VMBlock, outputting the enhanced joint features \( F_{JL}^{'} \) and \( F_{JR}^{'} \). Finally, the Dual Hand Parameter Regressor regresses the pose parameters (\( \theta_L \) and \( \theta_R \)) and shape parameters (\( \beta_L \) and \( \beta_R \)) of the MANO model, and through global average pooling and fully connected layers, it computes the 3D relative translation between the two hands.

\subsection{Backbone}
\label{Backbone}

We utilize the ResNet-50 \cite{he2016deep}, a general-purpose image feature extractor, to capture an initial feature map \( F \) from the input image \( I \in \mathbb{R}^{H \times W \times 3} \) of two hands. ResNet-50 has been pretrained on the large-scale ImageNet \cite{deng2009imagenet}, enabling effective extraction of high-level features from input images \( I \). This pretrained model serves as a robust starting point for feature extraction, ensuring the model can quickly recognize relevant patterns in the hand images. The extracted feature map \( F \) has dimensions \( h \times w \times C \), where \( h = H/32 \) and \( w = W/32 \), with \( C = 2048 \) channels, providing a rich representation of the image. This output serves as the initial feature representation of the input image, capturing complex visual patterns and structural information of the hands. Subsequently, the feature map \( F \) is further processed through two separate 1 × 1 convolutional layers. Each convolution is followed by batch normalization, which helps stabilize training by normalizing the output of each layer, and a ReLU activation function to introduce non-linearity and improve feature expressiveness. As a result, two distinct feature maps left-hand (\( F_L \)) and right-hand (\( F_R \)) features are obtained. Although the derived features \( F_L \) and \( F_R \) share the same spatial dimensions \( h \times w \) but have reduced channel dimensions \( c = C/4 \), which makes them more compact and efficient for subsequent processing steps. This reduction in the number of channels helps focus the model's attention on the most important features while reducing computational complexity. These optimized feature maps provide a refined and tailored representation of each hand, setting the stage for more precise 3D interacting hand mesh recovery in the following stages of the VM-BHINet pipeline.

\subsection{Vision Mamba Interaction Feature Extraction Block(VM-IFEBlock)}
\label{IFEM}

The Vision Mamba Interaction Feature Extraction Block(VM-IFEBlock) is a central component of our proposed method, designed to extract and process the interaction features between the left and right hands. Through multi-stage feature processing and deep interaction extraction, the VM-IFEBlock effectively addresses the challenges posed by the visual and physical differences between the two hands, overcoming the limitations of existing methods in terms of deep feature fusion. 

The Vision Mamba Block (VMBlock) and the Interaction Feature Extraction Module (IFEM) are the two main innovations of the VM-IFEBlock, playing a crucial role in extracting and processing the interaction features between the left and right hands. Specifically, VM-IFEBlock first utilizes an initial convolutional layer for dimensionality reduction and feature extraction. Then, the newly designed VMBlock performs complex transformations and interactions on the concatenated left and right hand features. Following this, the IFEM employs a non-local attention mechanism to establish strong relationships between the left and right hand features, further enhancing the model’s understanding of the deep interdependencies between the two hands. This is crucial for accurate hand reconstruction and understanding hand interactions.

\begin{figure}
\centering
\includegraphics[width=0.48\textwidth]{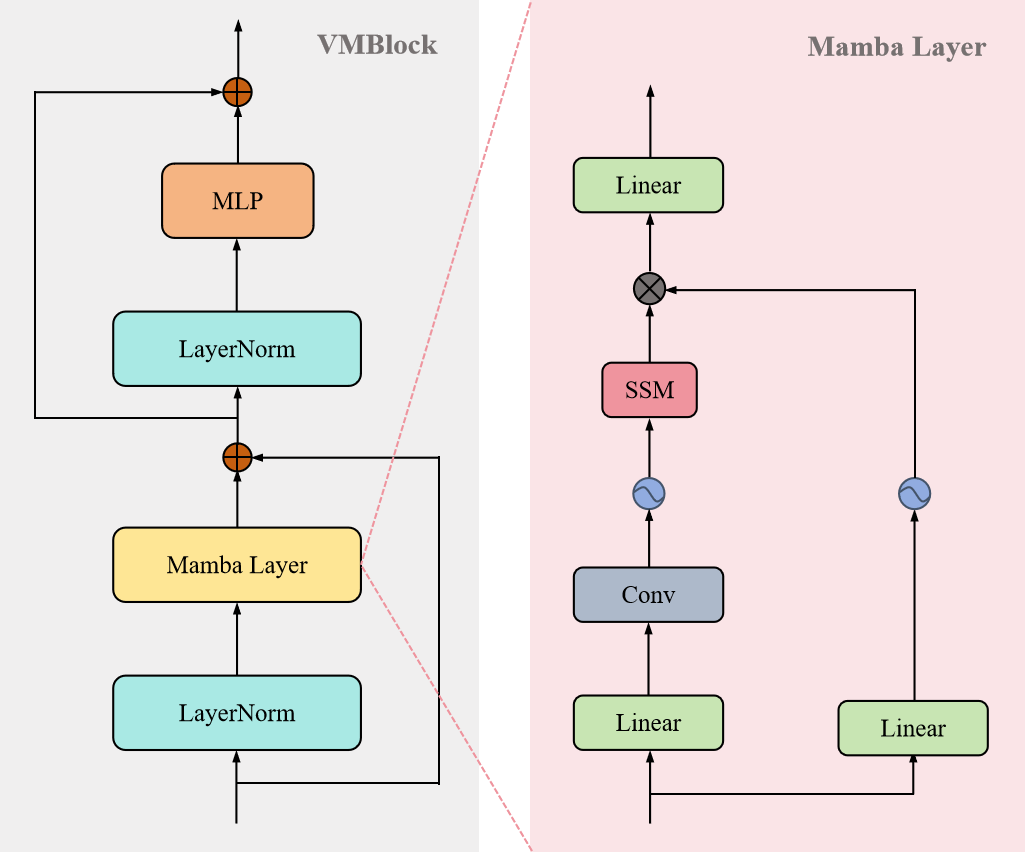}
\caption{The illustration of the proposed VMblock.}
\label{fig3}
\end{figure}

\textbf{Vision Mamba Block (VMBlock).} The Vision Mamba Block is the main construction block of the VM-IFEBlock, specifically designed to enhance the expressive capability of the left and right hand features and to process and capture the interaction features between the two hands. The Vision Mamba model is an advanced deep learning architecture specifically designed to handle complex visual tasks. It integrates various techniques to enhance feature expression and processing capabilities, enabling it to effectively capture and handle dynamic changes and improve the understanding of bimanual interactions. As illustrated in Figure \ref{fig3}, the Vision Mamba model combines components such as State Space Model (SSM), convolutional layers (Conv), Multi-Layer Perceptron (MLP), Layer Normalization (LayerNorm), and linear layers (Linear).

In Vision Mamba, the State Space Model in the Mamba layer is used to model dynamic systems and effectively capture temporal and spatial dependencies. Through the State Space Model, the model can handle long-term dependencies and dynamic changes in the input data, which is crucial for understanding the complex dynamics of hand poses and interactions. The State Space Model describes the dynamic behavior of the system using state transition and observation matrices, capturing relationships across different time steps and spatial locations. Convolutional layers are used to extract local features and capture detailed information from images, which is important for understanding subtle changes in hand poses. The convolution operation performs local perception on the input data through sliding windows, enabling effective extraction of spatial features. The MLP includes multiple fully connected layers and activation functions (ReLU) to further process and optimize features, enhancing the model's non-linear expression ability. Through the combination of multiple neurons, the MLP can learn complex feature representations. LayerNorm is used for normalization, improving the model's training stability and convergence speed. By normalizing the output of each layer, LayerNorm reduces internal covariate shift and accelerates the training process. The linear layers perform linear transformations, helping the model map features across different spaces. These transformations enhance the model’s expressive power, enabling it to better adapt to the task of reconstructing bimanual models in various complex environments. Additionally, the Vision Mamba model optimizes its computational process and reduces redundant computations, significantly lowering the computational burden and improving processing efficiency. This makes the model more efficient when handling high-dimensional features.

The following provides a detailed explanation of the innovations in VMBlock from multiple perspectives. The core innovation of VMBlock lies in the introduction of the State Space Model. State Space Model effectively models dynamic behaviors and captures temporal and spatial dependencies. In VMBlock, State Space Model performs multi-level processing across the global feature space, addressing the challenges posed by significant differences in hand poses and interactions. It establishes relationships between global features, capturing dynamic variations in input data (such as left and right hand features) across different scales. For instance, State Space Model helps capture dynamic changes in hand interactions, improving the model's understanding of hand movements and poses. The VMBlock combines traditional local convolution operations with State Space Model to extract useful local features, laying the foundation for 3D interacting hand mesh recovery. By integrating feature information from both hands, VMBlock generates a global feature representation that reflects overall hand actions and interaction patterns, rather than focusing solely on local details. This representation helps the model understand hand positioning in space and identify dynamic changes, such as the relative relationship between the left and right hands during complex hand interaction tasks. Another significant innovation of VMBlock is its optimization in computational efficiency. Traditional models, such as the ViT (Vision Transformer) block, often suffer from heavy computational loads and inefficiency, especially when processing high-dimensional features. However, VMBlock optimizes the computational process and reduces redundant calculations, significantly lowering the computational burden and enhancing processing efficiency.

\textbf{Interaction Feature Extraction Module (IFEM).} The Interaction Feature Extraction Module (IFEM) \cite{bi2024interhandnet} aims to enhance the model's understanding of the relationship between the left and right hand by extracting the interaction information between their features. The core idea of this module is to effectively integrate and process the left and right hand features through a non-local attention mechanism, allowing the model to capture the global dependencies between the two hands. Unlike traditional local convolution operations that focus only on local regions, the non-local mechanism considers all parts of the image or feature map, ignoring their spatial distances. This mechanism is crucial for the model to understand complex hand interactions, such as holding objects, or manipulating objects, where different parts of the hand influence each other. Non-local attention can capture such long-range dynamic changes. Since the left and right hands often have significant differences in posture, size, and movement, these differences may pose challenges for feature extraction and fusion. The IFEM focuses on the relationship between the two hands, mitigating the impact of these differences on model performance. This interaction modeling is critical for tasks involving object manipulation or complex gestures. The IFEM further enhances the model's performance by enabling deep interaction and feature fusion. In addition to capturing the spatial relationships between the two hands, it also understands the temporal changes in hand movements. This feature fusion improves the model’s ability to understand dynamic hand changes, optimizing subsequent 3D interacting hand mesh recovery. To conclude, the introduction of the IFEM enables the model to not only better understand the spatial positions of the hands but also effectively recognize the interactions between them during dynamic changes, thereby providing more accurate predictions in complex hand interaction tasks.

\subsection{Hand Joint Feature Extractor (HJFE)}
\label{Hand Joint Feature Extractor (HJFE)}
The Hand Joint Feature Extractor (HJFE) estimates and extracts joint features for both hands using 2.5D joint coordinates. First, it computes the 2.5D joint coordinates \( J_L \) and \( J_R \) by applying a \( 1 \times 1 \) convolution to the enhanced hand features \( F_{enh}^{L*} \) and \( F_{enh}^{R*} \), generating a 2.5D heatmap:

\begin{equation}
\hat{H}_L = \text{Conv}_{1 \times 1}(F_{enh}^{L*}), \quad \hat{H}_R = \text{Conv}_{1 \times 1}(F_{enh}^{R*}),
\end{equation}
where \( \hat{H}_L \) and \( \hat{H}_R \) are the generated heatmaps for the left and right hands, respectively. The soft-argmax operation is then applied to extract the precise joint coordinates for the left (\( J_L \)) and right hands (\( J_R \)):

\begin{equation}
J_L = \text{Soft-argmax}(\hat{H}_L), \quad J_R = \text{Soft-argmax}(\hat{H}_R).
\end{equation}

Next, for joint feature extraction, grid sampling is performed on the feature maps at the estimated 2.5D joint positions to extract the joint features. Let \( P_L \) and \( P_R \) represent the set of joint positions for the left and right hands, respectively. The joint features \( F_{JL} \) and \( F_{JR} \) are obtained by sampling the feature maps \( \hat{F}_L \) and \( \hat{F}_R \) at the positions \( P_L \) and \( P_R \):

\begin{equation}
\begin{aligned}
F_{JL} &= \text{GridSample}(F_{enh}^{L*}, P_L), \\
F_{JR} &= \text{GridSample}(F_{enh}^{R*}, P_R),
\end{aligned}
\end{equation}
where \( F_{JL} \) and \( F_{JR} \) are the left-hand and right-hand joint features, respectively. These joint features provide the global contextual information necessary for accurate 3D interacting hand mesh recovery.

\subsection{Joint Vision Mamba Block (JVMBlock)}
\label{VMBlock}

The Joint Vision Mamba Block (JVMBlock) optimizes the traditional self-attention mechanism by incorporating the State Space Model (SSM), leading to a significant reduction in computational complexity and improved efficiency. For each hand, it refines the left-hand joint features \( F_{JL} \) and the right-hand joint features \( F_{JR} \) through the VMBlock, which outputs the enhanced joint features \( F_{JL}^{'} \) and \( F_{JR}^{' } \). This refinement process can be expressed as:

\begin{equation}
F_{JL}^{'}, F_{JR}^{' } = \text{VMBlock}(F_{JL}, F_{JR}),
\end{equation}
where \( F_{JL} \) and \( F_{JR} \) are the initial joint feature representations for the left and right hands, and \( F_{JL}^{'} \) and \( F_{JR}^{' } \) are the refined features after passing through the VMBlock. 

Unlike the self-attention mechanism, which involves high time complexity of \( O(n^2) \), the SSM has a fixed computational cost that is independent of the sequence length, effectively reducing redundant computations. The State Space Model captures temporal dependencies through recursive updates of its internal states, reducing the number of parameters to be learned, thus improving parameter efficiency. Particularly in cases of rapid or complex hand pose changes, the VMBlock is more adept at capturing and modeling these dynamic features, ensuring the model remains efficient and accurate even during fast-changing hand movements. This enables JVMBlock to achieve both computational efficiency and accurate dynamic feature modeling when handling fast-changing hand movements and complex interactions.

\subsection{Dual Hand Parameter Regressor (DHPR)}
\label{Dual Hand Parameter Regressor(DHPR)}
The Dual Hand Parameter Regressor (DHPR) module is responsible for regressing both the pose parameters (\( \theta_L \) and \( \theta_R \)) and the shape parameters (\( \beta_L \) and \( \beta_R \)) of the MANO hand model for each hand. It starts by concatenating the enhanced hand joint features \( F_{JL}^{'} \) or \( F_{JR}^{'} \) with the 2.5D joint coordinates \( J_L \) or \( J_R \). The pose parameters \( \theta_L \) and \( \theta_R \) are predicted by passing these concatenated features through fully connected layers:

\begin{equation}
\theta_L = \text{FC}\left( \text{contact}(F_{JL}^{'}, J_L) \right),
\end{equation}
\begin{equation}
\theta_R = \text{FC}\left( \text{contact}(F_{JR}^{'}, J_R) \right).
\end{equation}

Next, the shape parameters \( \beta_L \) and \( \beta_R \) are obtained by forwarding the enhanced hand features \( F_{JL}^{'} \) and \( F_{JR}^{'} \) through a fully connected layer after global average pooling:

\begin{equation}
\beta_L = \text{FC}\left( \text{avgpool}(F_{JL}^{'}) \right), 
\end{equation}

\begin{equation}
\beta_R = \text{FC}\left( \text{avgpool}(F_{JR}^{'}) \right).
\end{equation}

Additionally, the DHPR module computes the 3D relative translation \( T_{\text{rel}} \) between the two hands using global average pooling and fully connected layers on the enhanced hand features \( F_{enh}^{L*} \) and \( F_{enh}^{R*} \) from both hands:

\begin{equation}
T_{\text{rel}} = \text{FC}\left( \text{avgpool}(F_{enh}^{L*}, F_{enh}^{R*}) \right).
\end{equation}

The final 3D hand meshes are then generated by feeding the predicted pose and shape parameters into the MANO layers:

\begin{equation}
\text{V}_L, \text{V}_R = \text{MANO}(\theta_L, \beta_L), \text{MANO}(\theta_R, \beta_R).
\end{equation}

\begin{table*}[t]
\setlength{\tabcolsep}{4pt}
\centering
\caption{Results of the ablation study on the impact of VM-IFEBlock and JVMBlock on model performance. This table presents the effects of different module combinations on parameter count (Params), computational complexity (GFLOPS), and prediction accuracy (MPVPE and MPJPE). MPVPE and MPJPE values are reported for single-hand, two-hand, and overall scenarios.}
\tiny
\def\arraystretch{1.4}
\resizebox{\linewidth}{!}{%
\begin{tabular}{c|cccc|c|c|ccc|ccc}
\toprule

& \multicolumn{4}{c|}{\textbf{Module}} & \multirow{2}{*}{\centering \textbf{Params (M)}} & \multirow{2}{*}{\centering \textbf{GFLOPS}} & \multicolumn{3}{c|}{\textbf{MPVPE}} & \multicolumn{3}{c}{\textbf{MPJPE}} \\ \cline{2-5}\cline{8-13}
\multirow{-2}{*}{\centering \textbf{Method}} & 
\multicolumn{2}{c}{\textbf{VM-IFEBlock}} & 
\multicolumn{2}{c|}{\textbf{JVMBlock}} & & & \textbf{Single} & \textbf{Two} & \textbf{All} & \textbf{Single} & \textbf{Two} & \textbf{All} \\ 
\midrule

\multirow{3}{*}{\textbf{Contrast}}
& \multicolumn{2}{c}{\ding{55}} & \multicolumn{2}{c|}{\ding{55}} & 136.22 & 28.49 & 5.61 & 6.87 & 6.16 & 5.28 & 6.17 & 5.73 \\
& \multicolumn{2}{c}{\ding{51}} & \multicolumn{2}{c|}{\ding{55}} & 61.15  & 13.85 & 5.17 & 7.04 & 5.99 & 4.89 & 6.36 & 5.62 \\
& \multicolumn{2}{c}{\ding{55}} & \multicolumn{2}{c|}{{\ding{51}}} & 112.07 & 27.61 & 5.26 & 7.14 & 6.04 & 4.97 & 6.43 & 5.70 \\ \midrule

\rowcolor{gray!15}\multicolumn{1}{c|}{\textbf{Ours}} & \multicolumn{2}{c}{\ding{51}} & \multicolumn{2}{c|}{\ding{51}}  & \textbf{36.99}  & \textbf{12.97} & \textbf{4.69} & \textbf{6.32} & \textbf{5.44} & \textbf{4.41} & \textbf{5.77} & \textbf{5.09} \\ \bottomrule
\end{tabular}
}
\label{table1}
\end{table*}

Finally, the model is trained by minimizing a loss function, which is a weighted sum of L1 distances between the estimated and ground truth values for the pose parameters, shape parameters, joint positions, 3D meshes, and 3D relative translation:

\begin{equation}
\begin{aligned}
\mathcal{L} = & \lambda_1 \cdot \| \theta_L - \theta_L^{\text{gt}} \|_1 + \lambda_2 \cdot \| \theta_R - \theta_R^{\text{gt}} \|_1 \\
& + \lambda_3 \cdot \| \beta_L - \beta_L^{\text{gt}} \|_1 +  \lambda_4 \cdot \| \beta_R - \beta_R^{\text{gt}} \|_1 \\
& + \lambda_5 \cdot \| J_L - J_L^{\text{gt}} \|_1 + \lambda_6 \cdot \| J_R - J_R^{\text{gt}} \|_1 \\
& + \lambda_7 \cdot \| \text{V}_L - \text{V}_L^{\text{gt}} \|_1 + \lambda_8 \cdot \| \text{V}_R - \text{V}_R^{\text{gt}} \|_1 \\
& + \lambda_9 \cdot \| T_{\text{rel}} - T_{\text{rel}}^{\text{gt}} \|_1,
\end{aligned}
\end{equation}
Where \( \theta_L^{\text{gt}}, \theta_R^{\text{gt}}, \beta_L^{\text{gt}}, \beta_R^{\text{gt}} \) represent the ground truth values for the pose and shape parameters, \( J_L^{\text{gt}}, J_R^{\text{gt}} \) are the ground truth joint positions, \( \text{V}_L^{\text{gt}}, \text{V}_R^{\text{gt}} \) are the ground truth 3D hand meshes, and \( T_{\text{rel}}^{\text{gt}} \) is the ground truth 3D relative translation. Additionally, \( \lambda_1, \lambda_2, \dots, \lambda_9 \) are the weights applied to each term in the loss function. 

This loss function ensures that the model learns to accurately predict the pose and shape parameters, as well as the 3D meshes and relative translation, by comparing the predicted values to the ground truth values during training.

\section{Experiments}
\subsection{Implementation details}
All experiments are implemented using PyTorch \cite{paszke2017automatic} with the Adam optimizer \cite{2014Adam}, and the batch size is set to 32 per GPU (trained on four RTX 4090 GPUs). For training our model on the InterHand2.6M \cite{moon2020interhand2,moon2022neuralannot} dataset, our model is trained for 30 epochs, with learning rate annealing at the 10th and 15th epochs, starting from an initial learning rate of \(1 \times 10^{-4}\).

\begin{figure*}[h]
    \centering
    \includegraphics[width=0.98\textwidth]{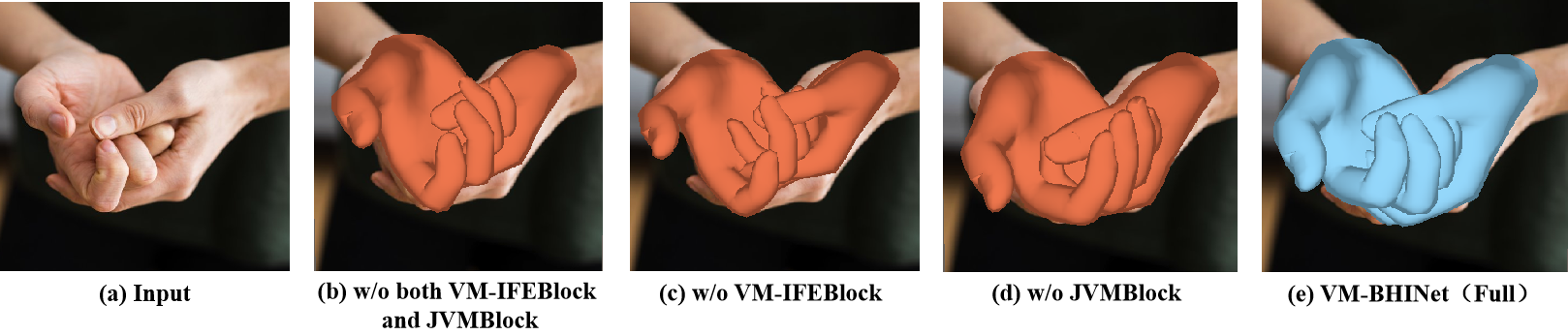}
    \caption{Qualitative Ablation Study on InterHand2.6M \cite{moon2020interhand2} dataset. The results show that our full model achieves the best performance compared to the versions that exclude certain components, where 'w/o' stands for 'without'.}
    \label{fig4}
\end{figure*}

\begin{table*}[t]
\caption{Quantitative comparison with state-of-the-art methods on InterHand2.6M \cite{moon2020interhand2} dataset. We present the results evaluated on the official test split of the InterHand2.6M dataset to ensure a fair comparison. * indicates that these results were obtained by evaluating the authors' released model on the official test split.}
\def\arraystretch{1.25}
\centering
\small
\begin{tabular}{c|c|c|ccc|ccc}
\toprule
\multirow{2}{*}{\textbf{Method}} & \multirow{2}{*}{\textbf{Params (M)}} & \multirow{2}{*}{\textbf{GFLOPS}} & \multicolumn{3}{c|}{\textbf{MPVPE}} & \multicolumn{3}{c}{\textbf{MPJPE}} \\
\cline{4-9}
& & & \textbf{Single} & \textbf{Two} & \textbf{All} & \textbf{Single} & \textbf{Two} & \textbf{All} \\ 
\hline
\multicolumn{9}{c}{\textbf{\textit{Model-free}}} \\  
\midrule
InterNet \cite{moon2020interhand2} & - & 19.49 & - & - & - & 12.16 & 16.02 & 14.22 \\
Kim et al. \cite{kim2021end} & - & - & - & - & - & - & - & 12.08 \\
DIGIT \cite{fan2021learning} & - & - & - & - & - & 11.32 & 15.57 & - \\
A2J-Transformer \cite{jiang2023a2j} & 42.00 & 25.65 & - & - & - & 8.10 & 10.96 & 9.63 \\
HDR \cite{meng20223d} & 55.00 & 15.47 & - & - & - & 8.51 & 13.12 & 10.97 \\
\hline
\multicolumn{9}{c}{\textbf{\textit{Model-based}}} \\ 
\midrule  
Keypoint Transformer \cite{hampali2022keypoint} & 117.05 & 19.66 & 10.16 & 14.36 & 11.94 & 10.99 & 14.34 & 12.78 \\
Two-Hand-Shape \cite{zhang2021interacting} & 143.37 & 28.98 & - & 13.95 & - & - & 13.48 & - \\
MeMaHand \cite{wang2023memahand} & - & - & - & - & 10.92 & - & - & 10.85 \\
IntagHand \cite{li2022interacting}* & 55.08 & 28.35 & 9.66 & 10.25 & 9.89 & 9.03 & 9.42 & 9.21 \\
ACR \cite{yu2023acr}* & 104.82 & 30.66 & 7.01 & 8.72 & 7.33 & 6.98 & 8.92 & 7.87 \\
EANet \cite{park2023extract}* & 136.22 & 28.49 & 5.61 & 6.87 & 6.16 & 5.28 & 6.17 & 5.73 \\
\rowcolor{gray!15}\textbf{VM-BHINet(Ours)} & \textbf{36.99} & \textbf{12.97} & \textbf{4.69} & \textbf{6.32} & \textbf{5.44} & \textbf{4.41} & \textbf{5.77} & \textbf{5.09} \\ 
\bottomrule
\end{tabular}
\label{table2}
\end{table*}

\subsection{Datasets and evaluation metrics}
\textbf{InterHand2.6M dataset.} The InterHand2.6M dataset \cite{moon2020interhand2,moon2022neuralannot} is the first large-scale, real-time, high-resolution 3D interacting hands dataset, designed to showcase the potential of 3D hand mesh recovery in interacting hands. Captured in a specially designed multi-camera studio environment, the InterHand2.6M dataset comprises 2.6 million frames of various single and strong interacting hand sequences from multiple subjects. It provides large-scale, diverse, and accurate GT 3D poses and meshes of interacting hands. The dataset was semi-automatically annotated and includes 1.36M training images and 849k test images. We trained and evaluated VM-BHINet on InterHand2.6M dataset.

\textbf{HIC dataset.} To demonstrate the versatility of our proposed VM-BHINet, we also evaluated its performance on the HIC \cite{tzionas2016capturing} dataset. The HIC dataset includes RGB-D sequences of single and interacting hands, along with 3D ground truth mesh models generated from 3D point clouds. Despite being recorded indoors, the dataset features diverse natural lighting and backgrounds, making it widely used for 3D hand mesh recovery in real-world settings.

\textbf{Evaluation metrics.} Mean per-joint position error (MPJPE) and Mean per-vertex position error (MPVPE) are key metrics for evaluating the accuracy of 3D hand pose and mesh reconstruction. MPJPE measures the Euclidean distance in millimeters between predicted and ground truth 3D joint positions, after aligning both hands by translating the root joint (e.g., wrist) to eliminate positional differences. It focuses on the precision of individual joint locations. MPVPE, on the other hand, evaluates the accuracy in millimeters of 3D mesh vertex positions, also using root joint alignment, and calculates the Euclidean distance between predicted and actual mesh vertices. MPVPE provides a more detailed assessment of the hand's surface shape reconstruction. Both metrics provide a comprehensive evaluation of hand pose and shape recovery by accounting for both joint and mesh vertex errors.

\begin{table*}[htbp]
\caption{Quantitative comparison with state-of-the-art methods on HIC \cite{tzionas2016capturing} dataset. We present the results evaluated on the official test split of the HIC dataset to ensure a fair comparison. * indicates that these results were obtained by evaluating the authors' released model on the official test split.}
\def\arraystretch{1.25}
\setlength{\tabcolsep}{12pt} 
\centering
\small
\begin{tabular}{c|ccc|ccc}
\toprule
\multirow{2}{*}{\textbf{Method}} & \multicolumn{3}{c|}{\textbf{MPVPE}} & \multicolumn{3}{c}{\textbf{MPJPE}} \\
\cline{2-7}
& \textbf{Single} & \textbf{Two} & \textbf{All} & \textbf{Single} & \textbf{Two} & \textbf{All} \\ 
\hline
\multicolumn{7}{c}{\textbf{\textit{Model-based}}} \\ 
\midrule
IntagHand \cite{li2022interacting}* & 47.78 & 49.31 & 48.13 & 46.87 & 47.19 & 46.90 \\
ACR \cite{yu2023acr}* & 46.02 & 40.32 & 39.10 & 45.38 & 42.98 & 47.09 \\
EANet \cite{park2023extract}* & 44.70 & 39.58 & 41.34 & 43.97 & 38.88 & 45.01 \\
\rowcolor{gray!15}\textbf{VM-BHINet(Ours)} & \textbf{27.72} & \textbf{29.17} & \textbf{28.98} & \textbf{26.82} & \textbf{28.66} & \textbf{27.47} \\ 
\bottomrule
\end{tabular}
\label{table3}
\end{table*}

\subsection{Ablation studies}

In order to verify the effectiveness of the proposed method VM-BHINet, an ablation study was conducted to evaluate the impact of different components on the model's performance. The study focused on two key components: the Vision Mamba Interaction Feature Extraction Block (VM-IFEBlock) and the Joint Vision Mamba Block (JVMBlock). These components were tested in isolation and in combination, comparing their effects on the performance metrics.

\begin{figure}
\centering
\includegraphics[width=0.49\textwidth]{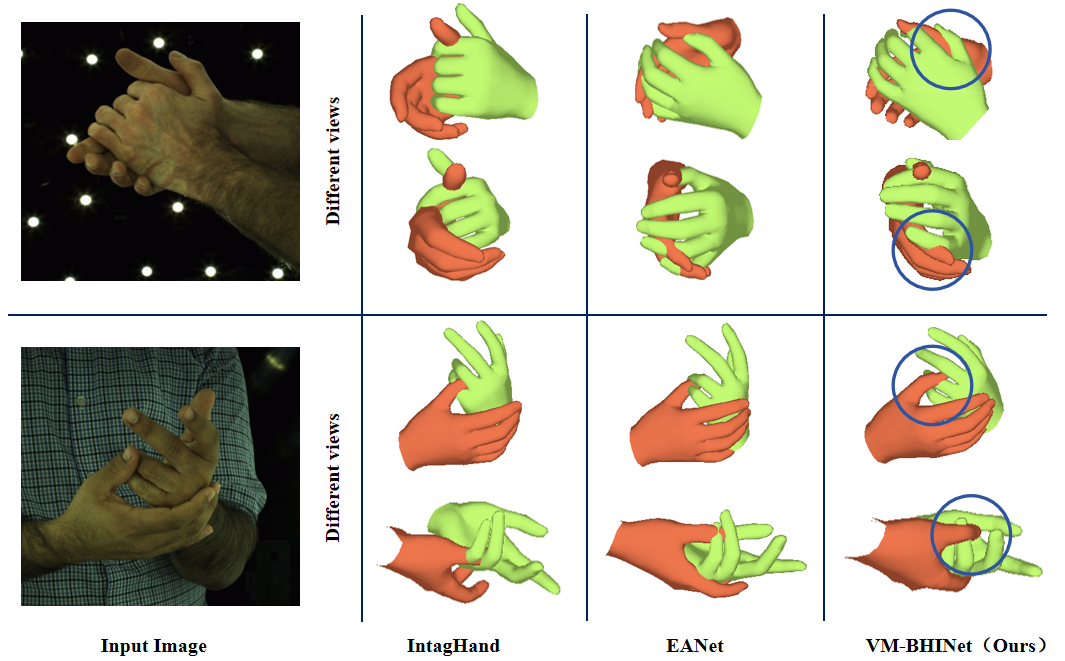}
\caption{Qualitative comparison of the interacting hand reconstruction with our method and the state-of-the-art methods IntagHand \cite{li2022interacting} and EANet \cite{park2023extract} on InterHand2.6M \cite{moon2020interhand2} dataset. Our approach demonstrates superior reconstruction quality across various viewpoints and different levels of interhand occlusion.}
\label{fig5}
\end{figure}

The results of the ablation study are as follows: when both VM-IFEBlock and JVMBlock were used together, the model achieved the lowest MPVPE and MPJPE values in 3D hand mesh reconstruction, indicating the best performance in accuracy. As shown in Table \ref{table1}, the model with both components demonstrated a significant improvement over the baseline model. In terms of parameter count and computational complexity, the combination of VM-IFEBlock and JVMBlock reduced the parameter count to 36.998 million and the GFLOPS to 12.974, which is much lower compared to using only VM-IFEBlock (61.150 million parameters and 13.854 GFLOPS) or JVMBlock (112.069 million parameters and 27.613 GFLOPS) independently. This reduction in both parameters and computational complexity highlights the efficiency of the proposed design. When only VM-IFEBlock was used, the MPVPE values for single-hand, two-hand, and overall scenarios were 5.17, 7.04, and 5.99, and the MPJPE values were 4.89, 6.36, and 5.62, respectively. Similarly, using only JVMBlock also led to performance improvement, with MPVPE values for single-hand, two-hand, and overall scenarios of 5.26, 7.14, and 6.04, and MPJPE values of 4.97, 6.43, and 5.70, respectively. However, the most significant performance improvement occurred when both components were used together, with MPVPE values of 4.69, 6.32, and 5.44, and MPJPE values of 4.41, 5.77, and 5.09 for single-hand, two-hand, and overall scenarios, respectively.

To complement the quantitative results, we conducted a qualitative ablation study on the InterHand2.6M dataset to illustrate the effects of each component in a more visual context. Figure \ref{fig4} illustrates the effect of removing each component, VM-IFEBlock and JVMBlock, and shows that the full model achieves the best visual accuracy in 3D hand mesh recovery, highlighting the advantages of using the full model over the individual components alone.

Overall, the ablation study demonstrates that the combination of VM-IFEBlock and JVMBlock not only enhances the model’s prediction accuracy but also significantly improves computational efficiency by reducing both parameter count and GFLOPS. The ablation study highlights the importance of both modules in improving the model's overall performance, confirming that their integration provides the best results in 3D hand mesh reconstruction tasks. These findings emphasize the effectiveness of the proposed approach in achieving high accuracy and efficiency simultaneously.

\subsection{Comparisons with state-of-the-art methods}

We present a quantitative comparison of our proposed VM-BHINet with state-of-the-art methods on the InterHand2.6M \cite{moon2020interhand2} and HIC \cite{tzionas2016capturing} dataset, as shown in Table \ref{table2} and Table \ref{table3}. Our method achieves competitive results across multiple evaluation metrics, including MPVPE and MPJPE.

In terms of MPVPE, our VM-BHINet performs excellently with a single-hand MPVPE of 4.69, two-hand MPVPE of 6.32, and overall MPVPE of 5.44 on the InterHand2.6M dataset, outperforming many existing methods such as Keypoint Transformer \cite{hampali2022keypoint}, Two-Hand-Shape \cite{zhang2021interacting}, IntagHand \cite{li2022interacting}, ACR \cite{yu2023acr} and EANet \cite{park2023extract}. Notably, our method significantly reduces the error when handling both single and two-hand interactions, highlighting the effectiveness of VM-BHINet in capturing fine-grained details of hand pose and interaction. Regarding MPJPE, our method also demonstrates strong performance, achieving 4.41 for single-hand, 5.77 for two-hand, and 5.09 for overall MPJPE, which is comparable to or better than other state-of-the-art methods, including EANet \cite{park2023extract}, IntagHand \cite{li2022interacting}, and A2J-Transformer \cite{jiang2023a2j}, and several others. These results confirm that VM-BHINet not only reduces the computational complexity but also ensures high accuracy in 3D hand mesh recovery.

Our VM-BHINet strikes the optimal balance between accuracy and computational efficiency, achieving significantly fewer parameters (36.99M) and lower GFLOPS (12.97). This makes it a highly promising solution for real-time applications that demand both fast and precise 3D hand mesh recovery. The results clearly demonstrate that VM-BHINet outperforms existing methods in both performance and efficiency, making it a state-of-the-art approach for interactive hand mesh recovery.

Furthermore, Figure \ref{fig5} presents visual comparisons with the previous state-of-the-art methods, IntagHand \cite{li2022interacting} and EANet \cite{park2023extract}, on the InterHand2.6M \cite{moon2020interhand2} dataset. The results showcase our VM-BHINet’s superior performance in accurately estimating hand poses and interactions.

\section{Conclusion and Future Work}
\textbf{Conclusion:} We present the Vision Mamba Bimanual Hand Interaction Network (VM-BHINet), a novel approach for 3D interacting hand mesh recovery. By incorporating state space models, our method effectively addresses challenges like occlusion and pose ambiguity in hand interactions. The core innovation, the Vision Mamba Interaction Feature Extraction Block (VM-IFEBlock), captures complex hand dependencies while optimizing computational efficiency. Experiments demonstrate that VM-BHINet outperforms state-of-the-art methods, achieving superior performance in both accuracy and efficiency on 3D Interacting Hand Mesh Recovery benchmarks.

\textbf{Limitation \& Future Work:} The current model's robustness under complex backgrounds and extreme lighting conditions needs improvement, and its real-time performance may require further optimization for practical applications. Additionally, the model's cross-domain generalization ability and capacity to handle multi-hand interaction scenarios still need enhancement. Future work could explore more lightweight architectures, self-supervised learning methods, and user feedback mechanisms to reduce reliance on labeled data and improve performance in complex environments. By addressing these challenges, VM-BHINet holds promise for broader applications in real-world scenarios.

\bibliographystyle{IEEEtran}
\bibliography{main}

\end{document}